\pdfoutput=1

\documentclass[11pt]{article}

\usepackage{acl}

\usepackage{times}
\usepackage{latexsym}

\usepackage[T1]{fontenc}

\usepackage[utf8]{inputenc}

\usepackage{microtype}
\usepackage{graphicx} 
\usepackage{comment}
\usepackage{multirow}
\usepackage{arydshln} 
\usepackage{subfigure}
\usepackage{amsmath} 
\usepackage{booktabs}

\title{Jointly Reinforced User Simulator and Task-oriented Dialog System with Simplified Generative Architecture}

\author{Hong Liu \and Zhijian Ou \\
  Speech Processing and Machine Intelligence (SPMI) Lab, Tsinghua University, Beijing, China\\
  \AND
  Yi Huang \and Junlan Feng\\
  China Mobile Research Institute, Beijing, China \\}
\newcommand{\modelname}{JRUD}
\newcommand{\Model}{SGA}
\begin{document}
\maketitle
\begin{abstract}
Recently, there has been progress in supervised funetuning pretrained GPT-2 to build end-to-end task-oriented dialog (TOD) systems.
However, online reinforcement learning of a GPT-2 based dialog system (DS), together with a end-to-end user simulator (US), has not ever been explored.
Moreover, a drawback with existing GPT-2 based TOD systems is that they mostly employ the whole dialog history as input, which brings inefficiencies in memory and compute.
In this paper, we first propose Simplified Generative Architectures (SGA) for DS and US respectively, both based on GPT-2 but using shortened history.
Then, we successfully develop Jointly Reinforced US and DS, called SGA-JRUD.
Our DS with the proposed SGA, when only supervised trained, achieves state-of-the-art performance on MultiWOZ2.1 and is more compute-efficient in both training and generation. 
Extensive experiments on MultiWOZ2.1 further show the superiority of SGA-JRUD in both offline and online evaluations.
\end{abstract}

\section{Introduction}
\label{sec:introduction}
\begin{figure}[t]
\centering
	\includegraphics[width=0.95\linewidth]{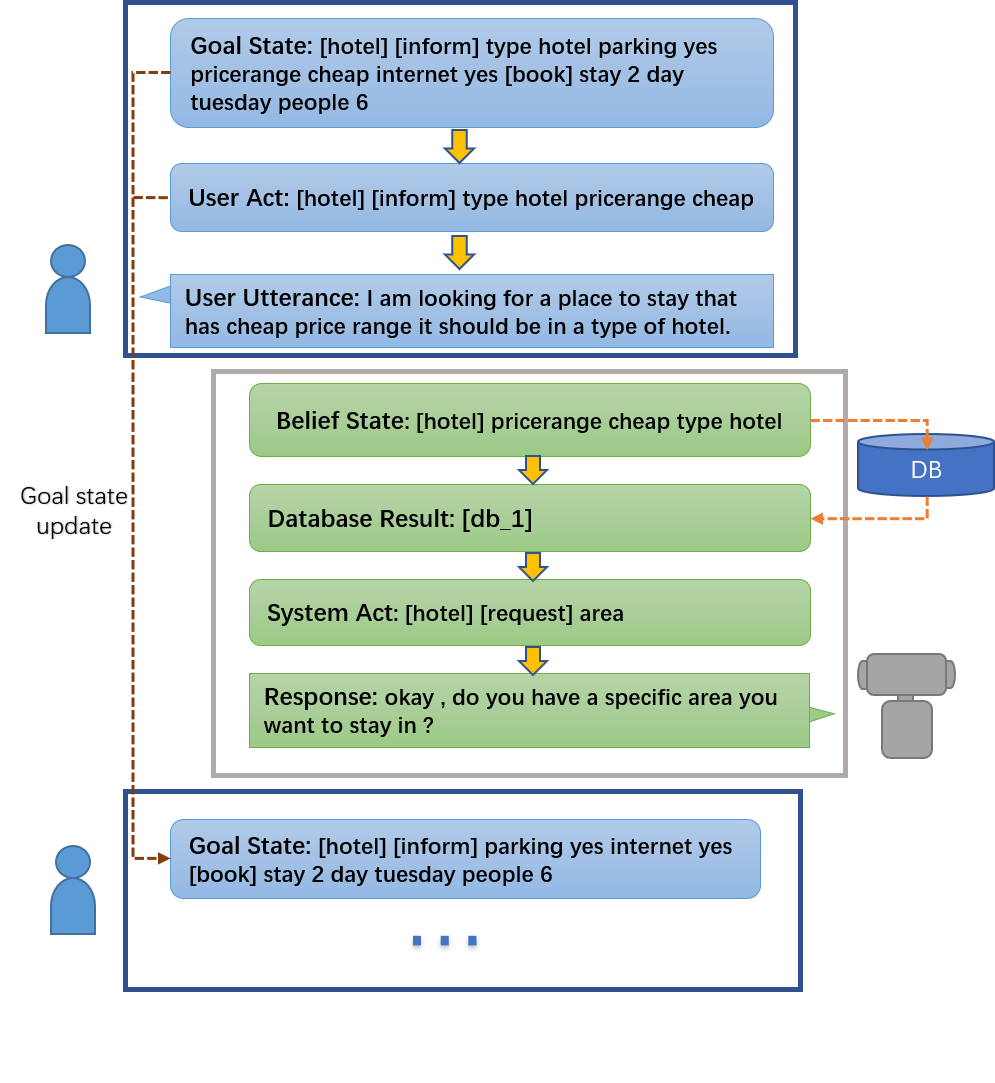}
	\vspace{-2em}
	\caption{The information flow in a task-oriented dialog. Square brackets denote special tokens in GPT-2.}
	\vspace{-1em}
	\label{fig:flow}
\end{figure}

Task-oriented dialog (TOD) systems, which are mainly designed to assist users to accomplish their goals, often consist of several modules including dialog state tracking (DST), database querying (DB), dialog policy (DP) and natural language generation (NLG). The information flow in a task-oriented dialog is illustrated in Figure~\ref{fig:flow}.
Recent studies recast these modules all as conditional generation of tokens and integrate them into a single language model (LM), which usually uses some pretrained language model (LM) such as GPT-2 \cite{radford2019gpt2} as the backbone.
Fine-tuning GPT-2 over annotated dialog datasets such as MultiWOZ \cite{budzianowski2018large} via supervised learning (SL) has shown state-of-the-art results \cite{hosseini2020simple, peng2020etal, kulhanek2021augpt, yang2021ubar}, thanks to the powerful generation ability of GPT-2. 

However, it has long been recognized that supervised learning over annotated dialog datasets alone may not be sufficient to learn a task-oriented dialog agent \cite{young2013pomdp}. Conversations often do not have only a single correct response, multiple responses can be appropriate for the same dialog context \cite{zhang2020task}. Supervised trained agents can become biased by the annotations.
Reinforcement learning (RL) for an agent aims to goal-directed learning from interaction between the decision-making agent and its environment \cite{sutton2018reinforcement} and is a natural choice for learning task-oriented dialog policies, where the user is modeled as the interactive environment.
Offline RL optimizes the policy from the fixed annotated dataset without online environment interaction \cite{zhou2017end, JEON2022101310} but only partially exploits the power of RL.
Online RL requires interaction with real humans or user simulators during training.
However, building a good user simulator is as challenging as designing a dialog agent, either rule based \cite{schatzmann-etal-2007-agenda} or data driven \cite{8639652, kreyssig-etal-2018-neural, tseng-etal-2021-transferable}.
There also have been some efforts to jointly optimize end-to-end dialog system (DS) and user simulator (US), but most are based on traditional architectures of using LSTM seq2seq networks \cite{DBLP:conf/asru/LiuL17,papangelis-etal-2019-collaborative,tseng-etal-2021-transferable}.

Inspired by the recent progress of funetuning pretrained LMs such as GPT-2 to develop the end-to-end trainable DS, in this paper we are firstly interested in building a GPT-2 based end-to-end trainable US for online RL of DS, which has not ever been explored.
Further, note that how to develop jointly optimized GPT-2 based DS and US in the RL framework is unclear, which requires new design of model architectures. Regarding this, we aim to develop \textbf{J}ointly \textbf{R}einforced \textbf{U}ser simulator and task-oriented \textbf{D}ialog system (JRUD), leveraging the recent progress of using pretrained LMs such as GPT-2 as the backbone.


To be clear, GPT-2 \cite{radford2019gpt2} in this paper refers to the particular class of causal LM, which computes conditional probabilities for next-token generation via self-attention based Transformer neural network \cite{vaswani2017attention}.
The basic idea in finetuning pretrained GPT-2 to build the dialog agent is to utilize the generation ability empowered by the finetuned causal LM.
Given a particular form of conditional model, $p(output|input)$, where $input$ and $output$ are token sequences, the GPT-2 LM can be finetuned over training samples $(input, output)$ (often referred to as training sequences \cite{hosseini2020simple}), and after finetuning, the model can be used for generation, i.e., generating $output$ after receiving $input$.

A limitation of previous methods in GPT-2 based DS, e.g., SimpleTOD \cite{hosseini2020simple}, SOLOIST \cite{peng2020etal}, AuGPT \cite{kulhanek2021augpt} and UBAR \cite{yang2021ubar}, is that the whole history is used as the input at each turn. 
This significantly increases the memory and computation cost in both training and generation.
Moreover, using the whole history may burden the model with redundant information and hurts the training efficiency.
To address the aforementioned limitation and to facilitate the development of JRUD, we propose \textbf{S}implified \textbf{G}enerative \textbf{A}rchitectures (\Model{}) for DS and US respectively, both based on GPT-2 but using shortened history.



The main contributions of this work can be summarised as follows:
\begin{itemize}
\item Our DS with the proposed SGA, called SGA-DS, when only supervised trained, achieves state-of-the-art performance on MultiWOZ2.1 \cite{eric2019multiwoz} and is more compute-efficient in both training and generation. 
\item To the best of our knowledge, our US with the proposed SGA, called SGA-US, represents the first GPT-2 based end-to-end trainable US, which could be trained via SL or RL.
\item Based on the proposed DS and US, we successfully develop a RL framework, called SGA-JRUD, for building jointly reinforced user simulator and dialog systems, which can be interacted and trained via online RL to significantly improve the performance of the TOD system, as shown in extensive experiments on MultiWOZ2.1.
\end{itemize}
\section{Related Work}
\label{sec:related}
\paragraph{End-to-end TOD systems} The methodology for building TOD systems is gradually advancing from separate training of individual modules \cite{mrkvsic2017neural, wen2017latent} to the end-to-end (E2E) trainable approach \cite{wen2017a, liu2017end, lei2018sequicity}. Recent studies have exploited the large-scale pre-trained language model such as GPT-2 for building end-to-end TOD systems, e.g., SimpleTOD \cite{hosseini2020simple}, SOLOIST \cite{peng2020etal},  AuGPT \cite{kulhanek2021augpt} and UBAR \cite{yang2021ubar}.
While existing GPT-2 based TOD systems achieve improved performance, these models mostly employ the whole dialog history as input during training and generation, which brings inefficiencies in computation, memory and learning. It is shown in Sec. \ref{sec:att_weights} that earlier history beyond the previous turn are in fact weakly attended to in next-token generation.
In contrast, the simplified architecture proposed in our SGA-DS only uses the belief state and system response of the previous turn for generating the response in current turn.

\paragraph{RL in TOD systems and user simulators} Reinforcement learning, which aims to train an agent towards maximizing long-term cumulative rewards from interactions between the agent and its environment, could be divided in two classes, offline and online \cite{sutton2018reinforcement}. Both classes have been applied in TOD systems.
Offline RL only optimizes the dialog agent over fixed collected data and thus avoids building user simulators \cite{zhou2017end, zhao-etal-2019-rethinking, JEON2022101310}. 
Online RL, instead, needs to design a user simulator (US) and let the dialog agent interact with the user simulator (acting as the environment) to generate new dialogs, over which the dialog agent can be further optimized.
A variety of user simulators have been studied, either rule based or data driven. 
A typical example of rule based US is the agenda-based user simulator (ABUS) \cite{schatzmann-etal-2007-agenda}.
In the data driven US approach, different models are proposed to train USs from data using different architectures, e.g. GRU seq2seq \cite{8639652} LSTM seq2seq \cite{kreyssig-etal-2018-neural}.
In this paper, motivated by the recent success of GPT-2 based DS, we propose a new GPT-2 based US and further design its simplified generative architecture.



\paragraph{Joint training of DS and US} There have been some studies to jointly optimize end-to-end DS and US, but most are based on traditional architectures of using LSTM seq2seq networks \cite{DBLP:conf/asru/LiuL17,papangelis-etal-2019-collaborative,tseng-etal-2021-transferable}. Earlier studies use template-based NLG module for both DS and US \cite{DBLP:conf/asru/LiuL17} and work in single domain such DSTC2 \cite{DBLP:conf/asru/LiuL17,papangelis-etal-2019-collaborative}. 
Progress has been made to use neural network based generation and work in multi-domain \cite{tseng-etal-2021-transferable}. 
Different RL algorithms have been attempted such as policy gradient and actor-critic in \cite{DBLP:conf/asru/LiuL17}, Q-learning in \cite{papangelis-etal-2019-collaborative}.
This paper represents a further advance in designing GPT-2 based DS and US with new simplified architectures.


\begin{figure}[t!]
	\centering
	\includegraphics[width=\columnwidth]{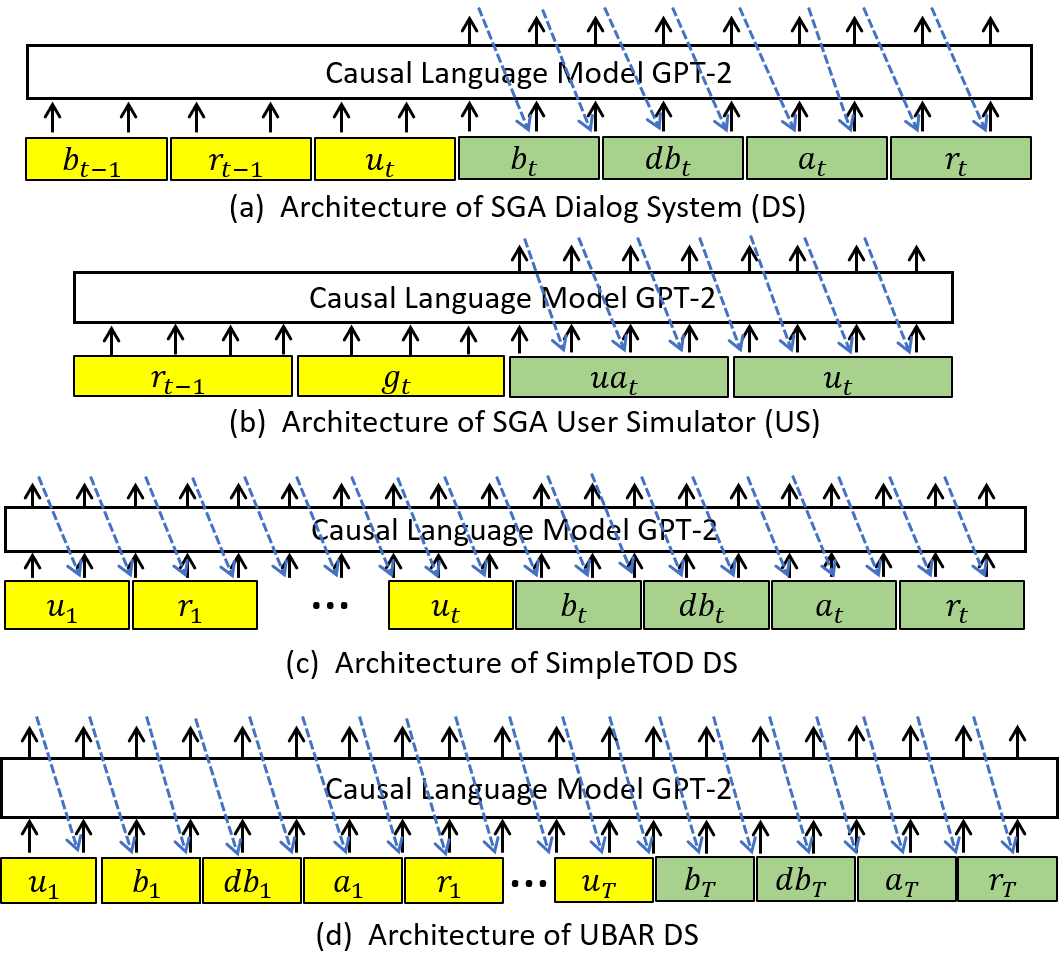}
	\caption{The proposed Simplified Generative Architectures (SGAs) for DS and US, shown in (a) and (b) respectively, as compared to SimpleTOD-DS (c) and UBAR-DS (d). Yellow boxes represent the conditioning $input$ of the model during generation, and green boxes the targeting $output$. The figure also reveals differences between our SGA models and the other two models. During supervised training, our SGA models are trained by maximizing the conditional likelihood of $output$ given $input$, while the other two models in fact maximizes the joint likelihood over both $input$ and $output$. Further, our SGA models can be naturally fit into the RL framework for DS and US respectively, while the other two models not (See Sec. \ref{sec:RL} for details).}
	\vspace{-0.8em}
	\label{fig:structure}
\end{figure}

\section{Method}
\label{sec:method}
In the following, we first introduce the background, then the simplified generative architectures (SGA) proposed for dialog system (DS) and user simulator (US), finally we describe the jointly reinforced method.
\subsection{Background}
\label{sec:background}
\paragraph{Notations}
According to the information flow in a task-oriented dialog as illustrated in Figure~\ref{fig:flow}, we let $g_t$ denote the user goal state, $ua_t$ the user act, $u_t$ the user utterance, $b_t$ the belief state, $db_t$ the database result, $a_t$ the system act and $r_t$ be the delexicalized response, respectively, at turn $t=1,\cdots,T$, for a dialog of $T$ turns.
In this work, all these variables are tokenized into token sequences, following recent studies in \cite{zhang-etal-2020-probabilistic, yang2021ubar}.
$\oplus$ denotes the concatenation of sequences such as in $u_t \oplus r_t$.
$|u_t|$ denotes the length of $u_t$ in tokens.
$\{u, r\}_t$ is a shorthand for $u_t, r_t$, and $\{u, r\}_{1:t}$ represents $\{u, r\}_1,\cdots,\{u, r\}_t$. 


\paragraph{Dialog system (DS)}
The main task for DS is, for each dialog turn $t$, to generate (or say, predict)\footnote{Note that database result $db_t$ is deterministically obtained by querying database using the predicted $b_t$. We omit $db_t$ in the discussion for simplicity.} $b_t$, $a_t$ and $r_t$, given $u_t$ and dialog history $u_1, r_1, \cdots, u_{t-1}, r_{t-1}$.
A recent progress in building DS is that all variables are represented by token sequences, and the workflow of a dialog system (belief state tracking, action and response generation) can be unified into a single sequence generation problem, which can be accomplished by a causal language model \cite{hosseini2020simple,yang2021ubar}.
Particularly, pretrained LM such as GPT-2 is finetuned to yield the conditional generation model $p(output|input)$, where $input$ and $output$ are token sequences with appropriate meanings. We can have different designs for $p(output|input)$, as long as it can perform the required prediction.
For example, SimpleTOD \cite{hosseini2020simple} generates according to a turn-level model $p(b_t, a_t, r_t|\{u, r\}_{1:t-1}, u_t)$, i.e., $\{u, r\}_{1:t-1}, u_t$ are concatenated as the $input$ and recursively generate the $output$, i.e., $b_t, a_t, r_t$, as shown in Fig. \ref{fig:structure}(c).
UBAR \cite{yang2021ubar} generates according to a session-level model, as illustrated in Fig. \ref{fig:structure}(d).


\subsection{Simplified Generative Architecture (SGA)}
\label{sec:structure}

A drawback with existing GPT-2 based TOD systems is that they mostly employ the whole dialog history as input during training and generation, as shown in Fig. \ref{fig:structure} for the examples of SimpleTOD and UBAR. This brings inefficiencies in memory, computation and learning.
It is shown later in Table 2 and Figure 3 that the memory cost in training and the time for running the DS to complete a dialog can be significantly reduced, if we use a shortened history as input in the DS model, while achieving state-of-the-art result on MultiWOZ2.1.
It is also shown in \cite{JEON2022101310} that the use of the whole dialogue history increases the training cost.
To address the aforementioned drawback and to facilitate the development of JRUD, we propose Simplified Generative Architectures (\Model{}) for both DS and US, as shown in Fig. \ref{fig:structure}(a) and (b) respectively, both based on GPT-2 but using shortened history.


\paragraph{SGA-DS} For DS to predict $b_t$, $a_t$ and $r_t$ at each turn $t$, we propose to use only the belief state $b_{t-1}$ and response $r_{t-1}$ from previous turn along with current user utterance $u_t$, instead of using the whole dialog history, as the conditioning input.
Presumably, this is reasonable since, by definition, belief state $b_{t-1}$ is generally a summary of dialog history up to turn $t-1$, which, together with $r_{t-1}$ and $u_{t}$ , should carry enough context information for the DS model to make prediction for $b_t$, $a_t$ and $r_t$.
Thus, we obtain our conditional model for DS, referred to as SGA-DS, which can be expressed as $p_{\theta}(b_t, a_t, r_t|b_{t-1}, r_{t-1}, u_t)$ and parameterized by $\theta$. 
In supervised learning, SGA-DS can be finetuned from pretrained GPT-2 by maximizing the following conditional likelihood:
\begin{equation}
\label{eq:sup-ds}
\begin{aligned}
    \mathcal{J}_{\text{DS-SL}} &=\log p_{\theta}(b_t, a_t, r_t|b_{t-1}, r_{t-1}, u_t)\\
    &= \sum_{i=1}^{|b_t \oplus a_t \oplus r_t|} \log p_\theta(c_i| b_{t-1}, r_{t-1}, u_t, c_{<i}) 
\end{aligned}
\end{equation}
where $c_i$ denotes the $i$-th token in $b_t \oplus a_t \oplus r_t$.

\paragraph{SGA-US} In the information flow shown in Fig.~\ref{fig:flow}, user goal state $g_t$ and user act $ua_t$ are introduced for building user simulator.
User goal refers to the predefined user task such as booking a cheap hotel, which is directly obtained from the annotation of the dataset; and the goal state represents the uncompleted part of the user goal. Both are represented by token sequences in this work.
The main task of US is to mimic an user, i.e., given the dialog history, to decide user act, generate user utterance, and update internal goal state to track progress towards satisfying the user goal.

In this work, we find that the approach of finetuning pretrained GPT-2 for conditional generation can be similarly applied to build US.
Particularly, for US to predict $ua_t$ and $u_t$ at each turn $t$, we propose to use the previous response $r_{t-1}$ and current goal state $g_t$, as the conditioning input.
The goal state $g_t$ is obtained by removing the slot values of previous user act $ua_{t-1}$ from previous goal state $g_{t-1}$ \footnote{\cite{tseng-etal-2021-transferable} adopts a similar goal state update, but uses binary vector to represent the goal state. Also note that this goal state update is deterministic, so we omit the generation of $g_t$ in Fig. \ref{fig:structure}(b) for simplicity.}. 
Thus, we obtain the conditional model for US, referred to as SGA-US, which can be expressed as $p_\phi(ua_t, u_t|r_{t-1}, g_t)$ and parameterized by $\phi$.
In supervised learning, SGA-US can be finetuned from pretrained GPT-2 by maximizing the following conditional likelihood:
\begin{equation}
\label{eq:sup-us}
\begin{aligned}
    \mathcal{J}_{\text{US-SL}} &=\log p_\phi(ua_t, u_t|r_{t-1}, g_t)\\
    &= \sum_{i=1}^{|ua_t \oplus u_t|} \log p_\phi(c'_i|r_{t-1}, g_t, c'_{<i})
\end{aligned}
\end{equation}
where $c'_i$ denotes the $i$-th token in $ua_t \oplus u_t$.

\subsection{Jointly reinforced US and DS (JRUD)}
\label{sec:RL}
After we design the architectures for DS and US, we can consider joint optimization of the two agents in the online RL framework.
The DS agent view the US as the environment and use its conditional model $p_{\theta}(b_t, a_t, r_t|b_{t-1}, r_{t-1}, u_t)$ as its policy; Conversely, the US agent view the DS as the environment and use its conditional model $p_\phi(ua_t, u_t|r_{t-1}, g_t)$ as its policy.
Here the policy of SGA-DS involves generating not only system act $a_t$, but also belief state $b_t$ and system response $r_t$.
This is different from some previous studies of learning reinforced DS, e.g., \cite{DBLP:conf/asru/LiuL17,papangelis-etal-2019-collaborative,tseng-etal-2021-transferable}, which only use RL to optimize the selection of system acts (but all use traditional architectures).
The action space of SGA-DS becomes larger, but thanks to the representation power of GPT-2, recursively predict (or say, decide about) $b_t$, $a_t$ and $r_t$ in one policy yields the best performance in our experiment.
In Sec~\ref{sec:policy}, we compare different schemes for policy definition for the DS agent with more discussions.

After supervised finetuning of DS and US separately, we apply the policy gradient method \cite{sutton2000policy} to jointly optimize the two agents.
We first let the two agents interact with each other based on the user goals sampled from training set and generate mini-batches of dialogs. Then we calculate the reward $R_t$ for each turn, which is described in detail in Sec~\ref{sec:reward}. The return $U_{i,t}$ for the action of turn $t$ at the $i$-th step is $\gamma^{|A_t|-i}R_t$, where $\gamma$ is the discounting factor and $|A_t|$ is the policy sequence length of turn $t$. We update the two agents with the following policy gradients:
\begin{align*}
    &\nabla_\theta \mathcal{J}_{\text{DS-RL}}=\sum_{i=1}^{|b_t \oplus a_t \oplus r_t|} U_{i,t} \nabla_\theta \log p_\theta(c_i)\\
    &\nabla_\phi \mathcal{J}_{\text{US-RL}}=\sum_{i=1}^{|ua_t \oplus u_t|} U_{i,t} \nabla_\phi \log p_\phi(c'_i)
\end{align*}
where $p_\theta(c_i)$ and $p_\phi(c_i)$ are shorthands for $p_\theta(c_i| b_{t-1}, r_{t-1}, u_t, c_{<i})$ and $p_\phi(c'_i|r_{t-1}, g_t, c'_{<i})$, respectively.

\section{Experiments}
\label{sec:exp}
\subsection{Dataset}
We use MultiWOZ2.1 \cite{eric2019multiwoz} for experiments. MultiWOZ2.1 is a large-scale English multi-domain task-oriented dialog datasets of human-human conversations. It contains 10.4k multi-turn dialogs, spanning over seven domains. 
In our experiments, we removed some inappropriate state values and corrected some spelling errors in the dataset, which is detailed in Appendix~\ref{sec:clean}.

\subsection{Evaluation}
Plenty of methods have been tested on MultiWOZ2.0 or MultiWOZ2.1, but may suffer from the inconsistencies in evaluation, which is analyzed in \citet{nekvinda-dusek-2021-shades}. To rigorously compare our model with others, we use their standardized evaluation scripts, which are now also the scripts adopted in the MultiWOZ website.
There are mainly four metrics for offline evaluation (corpus-based evaluation). \emph{Inform Rate} measures how often the entities provided by the system are correct. \emph{Success Rate} refers to how often the system is able to answer all the requested attributes by user. \emph{BLEU Score} is used to measure the fluency of the generated responses. And the \emph{Combined Score} is computed as (BLEU + 0.5 * (Inform + Success)). We also use the joint goal accuracy to evaluate DST performance, which is the proportion of dialog turns where all slot values are correctly predicted.
Noting that when performing online evaluation, i.e., evaluating the interaction quality of two agents, only Inform and Success can be calculated and the above scripts are no longer applicable, so we calculate Inform and Success rate using the scripts of \citet{tseng-etal-2021-transferable}.

To compare results, we conduct significance test for \emph{Success Rate}, \emph{Inform Rate}, and \emph{BLEU} using matched pairs test \cite{gillick1989some} and report the p-value.

\subsection{Training Procedure}
\label{sec:procedure}
We first train the DS and US separately on training set based on the SL objective described in Eq.~\ref{eq:sup-ds} and Eq.~\ref{eq:sup-us}. The resulting models are referred to as \Model{}-DS-SL and \Model{}-US-SL.
Then we conduct RL experiments through the interaction between the two agents. During interaction, we end the dialog according to the following conditions: 1) the number of dialog turns exceeds the threshold (20 in our experiment); 2) the goal state of US is empty; 3) the two agents generate ending intentions such as \verb|bye| and \verb|thank| concurrently.

To ameliorate the non-stationarity problem when jointly training the two agents \cite{DBLP:conf/asru/LiuL17},
we first fix the DS and optimize the US for 100 training cycles (each cycle contains 128 episodes) and obtain the model \Model{}-US-RL. Then we fix \Model{}-US-RL and optimize the DS for another 100 training cycles and obtain the model \Model{}-\modelname{}. 
In order to show the effect of joint optimization, we fix US and only optimize DS for 100 cycles and obtain \Model{}-DS-RL for comparison.
Remarkably, to avoid our dialog system deviating from natural language, we alternate RL updates with supervised learning at a certain ratio \cite{lewis-etal-2017-deal}, which is set to be 1:1 in our experiments.
More implementation details in our experiments are available in Appendix \ref{sec:implementation}.

\subsection{Supervised Benchmark Results}
\begin{figure}[t]
\begin{minipage}[t]{\linewidth}
    \includegraphics[width=0.99\linewidth]{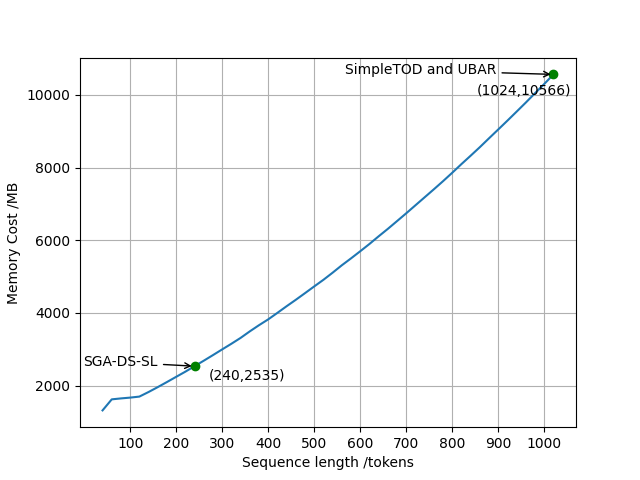}
	\captionsetup{font={small}}
	\caption{The memory costs during training with batch size 4, as a function of the lengths of training sequences.
	For \Model{}-DS-SL, SimpleTOD and UBAR, the means and standard deviations of the lengths of training sequences are 98$\pm$30, 190$\pm$112 and 440$\pm$220, respectively. The maximum sequence lengths for the three models are marked in the figure.}
	\label{fig:memory}
\end{minipage}
\end{figure}

We first show the offline evaluation results of different supervised trained DSs, which can be seen in Table~\ref{tab:baseline}. 
We evaluate different dialog systems in an end-to-end setting, which means that the generated belief states and system acts are used in response generation, i.e., the variables from the previous turn ($b_{t-1}$ and $r_{t-1}$) when used as the conditioning input are also the generated ones from the model itself.
\begin{table}[t]
\centering
\resizebox{\linewidth}{!}{
\begin{tabular}{lcccc}
\hline
\textbf{Model} &Inform &Success &BLEU &Combined\\
\hline
AuGPT \cite{kulhanek2021augpt} &76.6	&60.5 &16.8	&85.4\\
SOLOIST \cite{peng2020etal} 	&82.3	&72.4	&13.6 &90.9\\
UBAR \cite{yang2021ubar} &83.4	&70.3 &17.6	&94.4\\
\textbf{\Model{}-DS-SL} &84.90  &71.50  &18.14  &96.34\\
\hdashline
\textbf{\Model{}-DS-RL} &82.30  &70.70  &19.89  &96.39\\
\textbf{\Model{}-\modelname{}} &85.00  &74.00  &19.11  &98.61\\
\hline
\end{tabular}
}
\caption{Offline evaluation results on MultiWOZ2.1. Above the dashed line are supervised learning (SL) models and below are RL models.
The unbolded results are cited from the official website of MultiWOZ, which uses the same evaluation scripts of \citet{nekvinda-dusek-2021-shades} as in our experiments.}
\vspace{-0.5em}
\label{tab:baseline}
\end{table}
We can see that when only supervised trained, \Model{}-DS-SL achieves the highest combined score among all the GPT-2 based supervised DS models,
which indicates the superiority of our proposed \Model{} in building DS\footnote{Comparing SGA-DS-SL and UBAR, the p-values for Inform, Success and BLEU are 0.286, 0.477, 0.006, respectively. This shows that SGA-DS-SL achieves equally strong results as UBAR, not significantly better in all metrics, but being more compute-efficient, as shown in Figure ~\ref{fig:memory} and Table~\ref{tab:speed}.}.

Moreover, as shown in Figure~\ref{fig:memory} and Table~\ref{tab:speed} in Appendix~\ref{sec:time}, SGA-DS-SL is more compute-efficient than SimpleTOD and UBAR.
Since SGA-DS uses shortened history in training sequences, the training sequences for SGA-DS-SL are generally much shorter than for SimpleTOD and UBAR.
Consequently,  SGA-DS-SL consumes less training time and achieves faster generation speed.
These experiments are all conducted on a single 16GB Tesla-P100 GPU.


\subsection{RL Results}
\label{sec:rl_results}
To evaluate the RL experiments, we perform online evaluation where we let the two agents interact with each other for 1k times using the user goals of test set and calculate the inform and success rate of generated dialogs. Following \citet{shi-etal-2019-build}, we show the interaction results between different agents, which are shown in Table~\ref{tab:cross}.
\begin{table}[t]
\renewcommand\arraystretch{1.1}
\centering
\resizebox{\linewidth}{!}{
\begin{tabular}{llcc}
\hline
DS &US &Inform &Success\\
\hline
\multirow{2}{*}{\Model{}-DS-SL } &\Model{}-US-SL &87.0 &83.0\\
&\Model{}-DS-RL &89.0 &86.9\\
\hdashline
\multirow{2}{*}{\Model{}-DS-RL } &\Model{}-US-SL &90.0 &86.5\\
&\Model{}-US-RL &93.0 &90.1\\
\hdashline
\multirow{2}{*}{\Model{}-\modelname{}} &\Model{}-US-SL &87.0 &84.1\\
&\Model{}-US-RL &\textbf{95.0} &\textbf{92.9}\\
\hline
\end{tabular}
}
\caption{Online evaluation results between DSs and USs. Inform and Success rate are obtained by having the user simulator interacting with the dialog system on 1k user goals from the test corpus.}
\vspace{-0.5em}
\label{tab:cross}
\end{table}

In Table~\ref{tab:cross}, the inform and success rate increase significantly after RL. The jointly reinforced DS \Model{}-\modelname{} and \Model{}-US-RL achieve the highest success rate, which is higher than the supervised models (\Model{}-DS-SL and \Model{}-US-SL) by almost ten points. The substantial improvement implies that the new data generated during the interaction can really enhance our models. Moreover, we can see that \Model{}-\modelname{} obtains a higher success rate than \Model{}-DS-RL. This result indicates that the benefit of joint learning is that the US can improve its policy through the interaction between two agents, so that the DS learns better through the interaction with a better US. The improvement of the US is also reflected in the evaluation result between \Model{}-DS-RL and \Model{}-US-RL. The two models are never been trained together during RL, but they achieve pretty high success rate in the evaluation.

\begin{figure}[t]
	\includegraphics[width=0.95\linewidth]{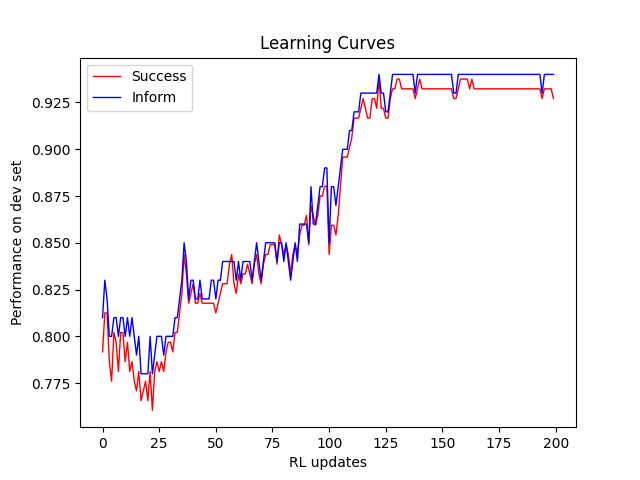}
	\caption{Inform and Success rate on the dev set during joint RL optimization.} 
	\label{fig:curve}
	\vspace{-0.8em}
\end{figure}

In Figure~\ref{fig:curve}, we also plot the learning curves of \Model{}-\modelname{}. In the first 100 training cycles, we fix DS and only optimize US. In the last 100 cycles, we fix the trained US and optimize DS only. We evaluate the two agents after every training cycle on the user goals of dev set. To reduce time cost, the dev set contains only 200 user goals randomly drawn from the original training and validation set before the running of RL. From Figure~\ref{fig:curve}, we can see that inform and success rate are both consistently improved during RL and finally converge to an upper bound in the second stage. 

Notably, it is observed in \citet{kottur-etal-2017-natural} that the improvement of success rate does not necessarily mean that the two agents can understand the semantic interaction, but may just invent an uninterpretable language. 
To address this concern, we conduct offline evaluation to see if our dialog system deviates from natural language after RL. The results are already included in Table~\ref{tab:baseline}. We can see that the BLEU scores do not decrease for reinforced DS models (SGA-DS-RL, SGA-JRUD). This indicates that our reinforced DS models can generate interpretable natural language after RL.

It can be also seen from Table~\ref{tab:baseline}, RL further improves the offline evaluation performance of \Model{}-DS in task completion. As shown in Table~\ref{tab:p-value}, the jointly reinforced model (SGA-JRUD) significantly improves over UBAR (p-value<0.02) and \Model{}-DS-SL (p-value<0.04) in Success Rate.


\begin{table}[t]
\centering
\resizebox{\linewidth}{!}{
\begin{tabular}{ccc}
\toprule
\Model{}-DS-SL vs UBAR &\Model{}-\modelname{} vs UBAR & \Model{}-\modelname{} vs \Model{}-DS-SL\\
\midrule
0.477 &0.015 & 0.030\\
\bottomrule
\end{tabular}
}
\caption{Significance test p-values for Success Rate between different models in offline evaluation.}
\label{tab:p-value}
\end{table}
\subsection{Analysis and Ablation Study}
\label{sec:ablation}
\subsubsection{Different reward settings}
\label{sec:reward}
A number of different settings for reward have been studied, as described in the following.\\
1) Success. If a dialog is successful, we set the reward of each turn to 1, otherwise it is set to be 0;\\
2) A turn-level synthetic reward similar to \citet{tseng-etal-2021-transferable, takanobu-etal-2020-multi}, which consists of requesting reward (+0.1 for each), repeating punishment (-0.5 for each) and global reward (proportions of tasks completed) of each agent;\\
3) A Sigmoid synthetic reward obtained by mapping the synthetic reward to [0,1] interval using Sigmoid function.

The above third setting is designed to exclude the influence of the value range of reward because the value range is different between the Success and the synthetic reward. It is found that using the first setting for award (i.e., 0 or 1 for each dialog according to Success) produces the best results in our experiments. All RL results in this paper are based on using this setting of reward, unless here for ablation study. 
The results of using different reward settings are reported in Table~\ref{tab:ablation1}.
\begin{table}[t]
\renewcommand\arraystretch{0.9}
\centering
\resizebox{0.65\linewidth}{!}{
\begin{tabular}{lcc}
\hline
Reward &Inform &Success\\
\hline
None &87.0 &83.0\\
\textbf{Success} &95.0 &92.9\\
Synthetic &94.0 &91.6\\
Synthetic-S &89.0 &86.4\\
\hline
\end{tabular}
}
\captionsetup{font={small}}
\caption{Online evaluation results in different reward settings. None denotes no RL, Synthetic denotes the synthetic reward, and Synthetic-S denotes the Sigmoid synthetic reward.}
\label{tab:ablation1}
\end{table}
We can see that all reward settings achieve better results than supervised baseline (Reward=None) and setting Success as reward achieves the best result. 

\subsubsection{Different policy schemes for DS}
\label{sec:policy}
The policy in RL refers to the probabilistic mapping from states to actions.
Previous studies of learning reinforced DS, e.g., \cite{DBLP:conf/asru/LiuL17,papangelis-etal-2019-collaborative,tseng-etal-2021-transferable}, mainly employ RL to optimize the DP module, i.e., use system acts for actions.
In contrast, the policy of SGA-DS involves generating not only system act $a_t$, but also belief state $b_t$ and system response $r_t$.
To compare policy schemes for reinforced DS, we conduct two other RL experiments, where the policy includes $a_t$ and $a_t \oplus r_t$  respectively.
We show the online and offline evaluation results in Table~\ref{tab:policy}.
\begin{table}
\centering
\resizebox{\linewidth}{!}{
\begin{tabular}{ccccc}
\hline
\multirow{2}{*}{Policy} &\multicolumn{2}{c}{Online Evaluation} &\multicolumn{2}{c}{Offline Evaluation}\\
\cmidrule(lr){2-3} \cmidrule(lr){4-5}
&Inform &Success &Combined &JointGoal\\
\hline
$b_t \oplus a_t \oplus r_t$ &95.0 &92.9 &98.61 &54.7\\
$a_t$ &93.0 &90.6 &99.23 &54.2\\
$a_t \oplus r_t$ &92.0 &89.9 &98.11 &54.3\\
\hline
\end{tabular}
}
\caption{The comparison of different optimization objective. We show both online and offline evaluation results.}
\label{tab:policy}
\end{table}

It can be seen from Table~\ref{tab:policy} that using $b_t \oplus a_t \oplus r_t$ for policy achieves the highest online evaluation results with large margins. In offline evaluation, using $b_t \oplus a_t \oplus r_t$ is also among the best. Using $a_t$ achieves higher combined score, but the  difference is not significant (p-value=0.355).
We provide two points, which may explain the advantage of our model in using $b_t \oplus a_t \oplus r_t$ for RL.
First, since the DST, DP and NLG modules in GPT-2 based DS share the model parameters, parameter adjust in one module will affect other modules. 
Only optimizing DP with RL without considering other modules may mislead other modules.
Using $b_t \oplus a_t \oplus r_t$ leads to better overall optimization and decision-making.
Second, the conflict between policy learning and NLG, which was a concern in previous studies when using modular or small-capacity architectures \cite{zhao-etal-2019-rethinking}, could be relieved, thanks to the high-capacity of GPT-2.




\subsection{Attention Weight Statistics} 
\label{sec:att_weights}
\begin{figure}[t]
	\centering
	
\begin{minipage}[t]{0.49\linewidth}
	\subfigure[UBAR]
	{\label{fig:attention_ubar}
	\includegraphics[width=\linewidth]{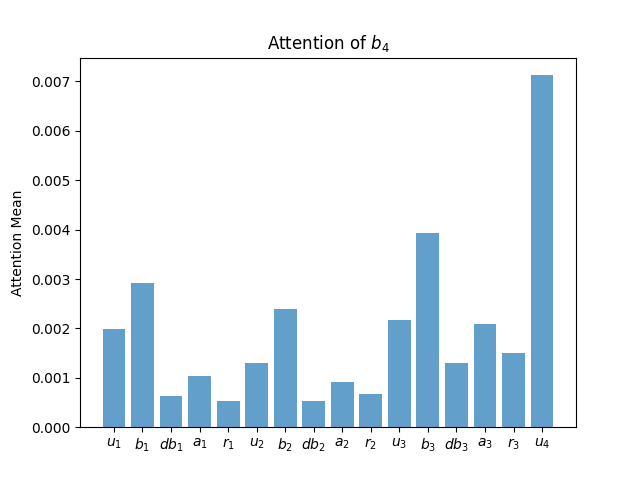}}
\end{minipage}
\begin{minipage}[t]{0.49\linewidth}
	\subfigure[\Model{}-DS-SL]
	{\label{fig:attention_jrud}
	\includegraphics[width=\linewidth]{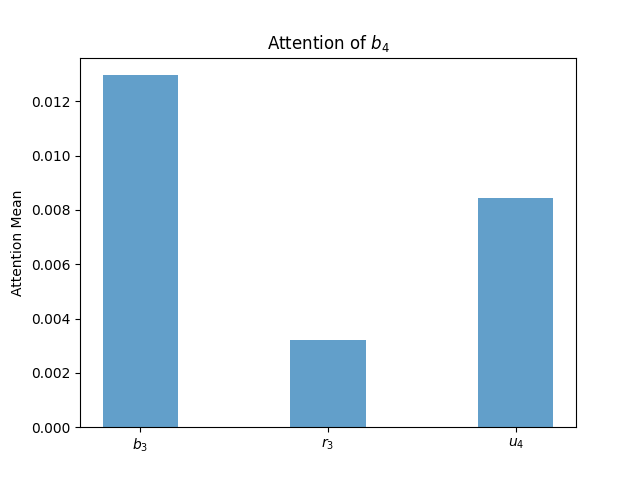} }
\end{minipage}
	\caption{Average attention weights for predicting the belief state in the 4-th turn.}
	\label{fig:attention}
	\vspace{-0.5em}
\end{figure}

In SGA, we propose that $b_{t-1}$, $r_{t-1}$ and $u_t$ could be sufficient for the DS to generate $b_t$, $a_t$ and $r_t$, and the whole dialog history contains redundancy. To support this idea, we calculate the average attention weights for prediction at a certain turn $t$ which point to the variables in previous all turns, using the session-level model UBAR \cite{yang2021ubar}.
Let $t=4$ and we show the means of the attention weights in generating $b_4$ in Figure~\ref{fig:attention_ubar}.

We can see that in UBAR, belief state $b_4$ mainly attends to current user utterance ($u_4$) and belief states of all previous turns ($b_1$, $b_2$ and $b_3$). 
Note that belief state is defined as the accumulation of history information, which means that $b_3$ contains almost all the slots and values of $b_1$ and $b_2$. Thus, the attentions to $b_1$ and $b_2$ are redundant, they appear mainly because there are no mechanism to reduce such redundancy in previous models.
In Appendix~\ref{sec:example-attn} , we provide an example to show how UBAR attends to previous belief states and we can see that the tokens in $b_1$ and $b_2$ with large attention weights are almost all appeared in $b_3$.
If we do not let the model attend to $b_1$ and $b_2$, the model will naturally attend more to $b_3$ and still not miss the information in $b_1$ and $b_2$. This can be indeed seen from the attention weights of our proposed SGA-DS, as shown in Figure~\ref{fig:attention_jrud}. Different from UBAR, \Model{}-DS-SL attends more to $b_3$ than to $u_4$ when generating $b_4$. In UBAR, the attentions are scattered across $b_1,b_2,b_3$.
We also show the means of attention weights in predicting system act and response in Appendix~\ref{sec:attn-appendix}. As expected, the attentions mainly point to the variables of current turn, especially the history variables closely nearby.


\section{Discussion and Conclusion}
In this paper, we first propose Simplified Generative Architectures (SGA) for DS and US respectively, both based on GPT-2 but using shortened history. Then, we successfully develop Jointly Reinforced US and DS, called SGA-JRUD.
The supervised trained DS with the proposed SGA achieves state-of-the-art performance on MultiWOZ2.1 and is more compute-efficient in both training and generation.
To develop and demonstrate JRUD, extensive experiments on MultiWOZ2.1 are conducted with both offline and online evaluations, we study different reward settings, different policy schemes. More discussions are provided in Appendix~\ref{sec:explore_action}, \ref{sec:example-improve}, \ref{sec:example-dial} on exploration of actions, example of improvement, examples of generated dialogs about our models, respectively.

This work represents a new step towards jointly reinforced end-to-end US and DS, but its performance may be limited by the pretrained GPT-2 backbone and the policy gradient algorithm used. Attempting larger backbone and new RL algorithms are interesting futher directions.
\bibliography{anthology,custom}
\bibliographystyle{acl_natbib}
\clearpage
\appendix

\section{Dataset Cleaning}
\label{sec:clean}
The details of how we clean up the dataset can be seen in Table~\ref{tab:correction}.
\begin{table}[t]
\centering
\resizebox{\linewidth}{!}{
\begin{tabular}{llc}
\hline
Correction &Example &Number\\
\hline
\multirow{2}{*}{Belief state} &[hotel] stars 4 internet yes name cambridge belfry &\multirow{2}{*}{24091}\\
&$\rightarrow$ [hotel] stars 4 internet yes &\\
Spelling &portugese $\rightarrow$ portuguese &415\\
\hline
\end{tabular}
}
\caption{Correction of the MultiWOZ2.1.}
\vspace{-0.5em}
\label{tab:correction}
\end{table}
Specifically, when training with delexicalized responses, the belief states of some turns become incorrect because they contain a redundant slot \verb|name| and its corresponding value. 
These slots and values are originally from some lexicalized responses in the dataset, but after delexicalization, the corresponding values in the responses are replaced with placeholders, which means that the TOD system needs to infer many names never appeared in the dialog history if we do not correct the belief states. The correction method is to simply delete some \verb|name| slots in belief states whose values never appeared in user utterances and delexicalized responses of previous turns. This correction will not have any bad impact on the TOD system, because the system can find the deleted names by querying the database with the remaining slot values in the belief state.
Another change is to correct some spelling errors of the word \verb|portuguese| in belief states and user utterances. 

\section{Implementation Details}
\label{sec:implementation}
We implement the models with Huggingface Transformers repository of version 3.5.1.
We initialize \Model{}-DS and \Model{}-US with DistilGPT-2 \cite{sanh2019distilbert}, a distilled version of GPT-2. During supervised pre-training, we use AdamW optimizer and a linear scheduler with 20\% warm up steps and maximum learning rate $1e^{-4}$. The minibatch base size is set to be 8 with gradient accumulation steps of 4.
The total epochs are 50 and we monitor the performance on validation set and apply early stopping (stop when the current best model keeps the best for subsequent 4 epochs). We select the best model on the validation set then evaluate it on test set.
During RL, we no longer use scheduler and fix the learning rate to $2e^{-5}$. The batch size is set to be 16 with gradient accumulation steps of 12. As described in Sec~\ref{sec:procedure}, we first optimize US for 100 training cycles and select the US with best online evaluation result. Then we optimize DS for another 100 training cycles with the selected US and select the best DS. Thanks to our proposed simplified structure, all the experiments above can be performed on a single 16GB Tesla-P100 GPU.

\section{Analysis and Case Study}
\subsection{Time Costs}
We use re-implementation of SimpleTOD and UBAR under the same optimizer and scheduler as SGA-DS-SL; 
The minibatch base sizes and gradient accumulation steps are (8,4) for SGA-DS-SL, and (2, 16) for SimpleTOD and UBAR, respectively. 
We monitor the performance on validation set and apply early stopping (stop when the current best model keeps the best for subsequent 4 epochs).
\label{sec:time}
\begin{table}[t]
\centering
\resizebox{0.7\linewidth}{!}{
\begin{tabular}{lcc}
\toprule
Model & Training Time & Generation Time\\
\midrule
\Model{}-DS-SL &204min &229s\\
SimpleTOD &396min &276s\\
UBAR &369min &352s\\
\bottomrule
\end{tabular}
}
\caption{Time costs of different models. The right two columns report the training time and the time of generation on test set respectively.}
\label{tab:speed}
\end{table}
\subsection{Attention Weights}
\label{sec:attn-appendix}
\begin{figure}[t]
	\centering
	\subfigure[Average attention weights for system act]
	{\label{fig:attention_a}
	\includegraphics[width=\columnwidth]{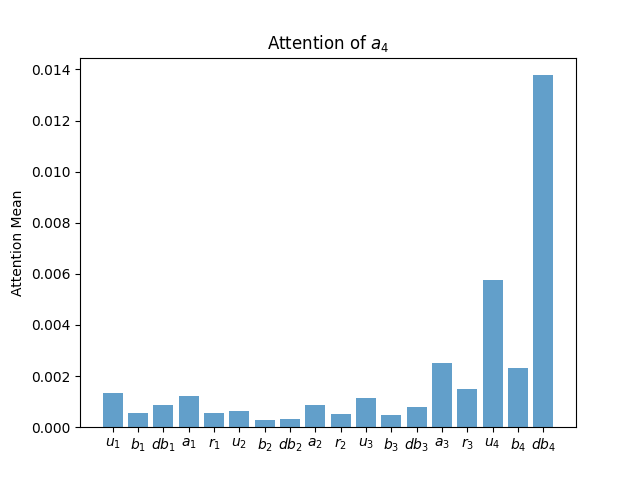}}
	\subfigure[Average attention weights for response]
	{\label{fig:attention_r}
	\includegraphics[width=\columnwidth]{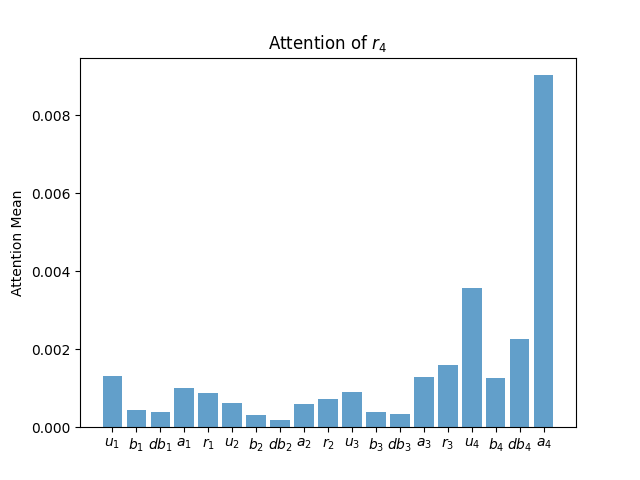} }
	\caption{Average attention weights for predicting the system act and response in UBAR.}
	\label{fig:attention-append}
\end{figure}

We show the statistical results for system act and response in UBAR, which are shown in Figure~\ref{fig:attention-append}. It can be observed that the two variables almost only attend to the variables of current turn, especially the history variables closely nearby. For instance, system act $a_t$ most attends to the database result $db_t$ and response $r_t$ most attends to the system act $a_t$. This is reasonable because TOD systems always make decisions based on what the system finds in the database and generate a response that is highly consistent with the selected action.
\subsection{Example of UBAR's Attention} 
\label{sec:example-attn}
\begin{figure}[t]
    \subfigure[Attention weights of belief state]
	{\label{fig:heatmap_b}
	\includegraphics[width=\linewidth]{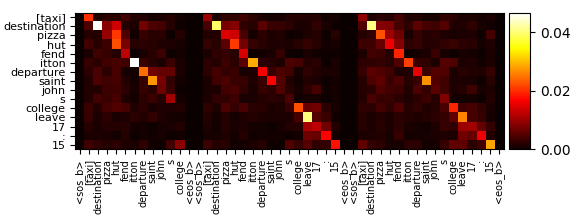}
	}
	\subfigure[Attention weights of user utterance]
	{\label{fig:heatmap_u}
	\includegraphics[width=\linewidth]{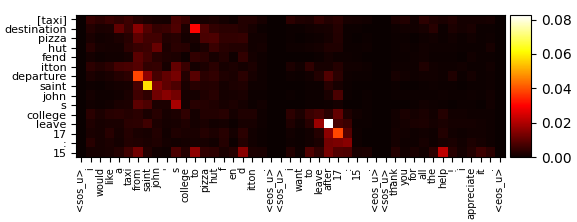}
	}
	\caption{The heat map of attentions. The vertical axis represents the belief state of the fourth turn, the horizontal axis represents the belief state or user utterance of previous turns ($b_{1:3}$ or $u_{1:3}$ are merged together). }
	\label{fig:heat}
\end{figure}

We provide an example of a training sequence in Figure~\ref{fig:heat} to show how the belief state of 4-th turn in UBAR \cite{yang2021ubar} attends to belief states and user utterances of the previous three turns.
As we can see, each token of belief state in the fourth turn basically gives the same token the maximum attention weight in previous belief states. But the tokens of the first two belief states are included in the third belief state, which indicates that if we constrain the model to only attend to the third belief state, it can also absorb enough information.
\subsection{Exploration of Actions}
\label{sec:explore_action}
In order to investigate whether our model has the ability to generate richer dialog actions after RL, we count the number of actions that have not appeared in training set during evaluation (unseen acts). 
The statistics can be seen in Table~\ref{tab:exploration}.
\begin{table}[t]
\centering
\resizebox{\linewidth}{!}{
\begin{tabular}{ccc}
\toprule
Model & Unseen System Acts & Unseen User Acts\\
\midrule
\Model{}-DS-SL &62 &6\\
\Model{}-\modelname{} &74 &8\\
\bottomrule
\end{tabular}
}
\caption{Statistics on dialog acts.}
\label{tab:exploration}
\end{table}
It can be seen that the number of both unseen system acts and unseen user acts has increased after RL. RL does improve the richness of dialog actions to a certain extent. However, the total unseen system acts in the original test set is 556, much larger than the number of generated unseen system acts. This means that although the diversity of dialog actions has improved, it is not enough and needs further study.

\subsection{Example of improvement from RL}
\label{sec:example-improve}
To help understanding the improvement from RL, we show some examples in Table~\ref{tab:case}. We selected two dialogs in test set and compare the system act and delexicalized response generated by the supervised model \Model{}-DS-SL and the RL model \Model{}-\modelname{}. 

From the first example, we can see that \Model{}-DS-SL forgets to inform some important attributes such as address, while \Model{}-\modelname{} tends to generate more attributes to accomplish tasks.
In the second example, \Model{}-DS-SL finishes booking when the user has not provided enough information, which leads to a wrong response in the next turn. However, \Model{}-\modelname{} learns to ask the user more information when the booking cannot be completed, and to complete the booking once the information is enough. 

We also provide some dialogs generated during the interaction between the US and DS agents, which can be seen in Appendix~\ref{sec:example-dial}.
\begin{table}[ht]
\centering
\resizebox{\linewidth}{!}{
\begin{tabular}{ll}
\toprule[1.2pt]
\multicolumn{2}{c}{\textbf{SNG1086}}\\
\hline\hline
\multirow{2}{*}{User} &i am looking for some info on an attraction called \\
     &kettle's yard.\\
Act &[attraction] [inform] price area address\\
\multirow{2}{*}{Resp} &absolutely! the entrance is [price]! it is located\\
&in the [area]. the address is [address]!\\
\hline
Act$_{sup}$ & [attraction] [inform] price area\\
Resp$_{sup}$ & sure! it is located in the [area] and costs [price].\\
\hline
Act$_{RL}$ &[attraction] [inform] price area name type address\\
\multirow{2}{*}{Resp$_{RL}$} &[name] is a [type] located in the [area] at [address].\\
&the entrance fee is [price].\\
\midrule[1.2pt]
\multicolumn{2}{c}{\textbf{SNG01797}}\\
\hline\hline
User &i'll be staying for 3 nights starting on tuesday.\\
Act &[hotel] [request] people\\
Resp &for how many guests please, sir?\\
\hline
Act$_{sup}$ &[hotel] [offerbooked] reference\\
\multirow{2}{*}{Resp$_{sup}$} &booking was successful. reference number is: \\
&[reference].\\
\hline
Act$_{RL}$ &[hotel] [request] people\\
Resp$_{RL}$ &how many people will be staying?\\
\hline\hline
User &1 guest for 3 nights\\
Act &[hotel] [offerbooked] reference\\
\multirow{2}{*}{Resp} &booking was successful. reference number is: \\
&[reference].\\
\hline
Act$_{sup}$ &[hotel] [request] day\\
Resp$_{sup}$ &what day would you like to start your stay?\\
\hline
Act$_{RL}$ &[hotel] [offerbooked] reference [general] [reqmore]\\
\multirow{2}{*}{Resp$_{RL}$} &booking was successful. reference number is:\\
&[reference]. is there anything else i can help you?\\
\bottomrule[1.2pt]
\end{tabular}
}
\caption{Two dialog examples (only 1 or 2 turns are shown) selected from test set. Act and Resp denote the system act and response from dataset. Act or Resp with the subscript $sup$ and $RL$ denote they are generated by \Model{}-DS-SL and \Model{}-\modelname{}.}
\label{tab:case}
\end{table}

\subsection{Example of Generated Dialogs}
\label{sec:example-dial}
We provide a dialog sample generated during the interaction between \Model{}-\modelname{} and \Model{}-US-RL in Table~\ref{tab:interaction-case}. We can see that there's a big difference between the generated dialog and the original dialog, very conducive to model learning.
\begin{table}[ht]
\centering
\resizebox{\linewidth}{!}{
\begin{tabular}{ll}
\toprule[1.2pt]
\multicolumn{2}{c}{\textbf{SNG1076}}\\
\hline\hline
\multirow{2}{*}{User} &Can you see if there are any entertainment attractions in the\\
&centre of town? \\
\specialrule{0em}{0.2em}{0.2em}
\multirow{2}{*}{Resp} &Unfortunately there are no [type] attractions in the [area]. Is\\
&there a different area of town that would work for you ?\\
\specialrule{0em}{0.2em}{0.2em}
User$_{gen}$ &I am looking for a museum in the centre of town.\\
\specialrule{0em}{0.2em}{0.2em}
\multirow{2}{*}{Resp$_{gen}$} &There are [choice] [type] in the centre of town. Is there a\\
&particular price range you are looking for?\\
\hline
User &What about a museum?\\
\specialrule{0em}{0.2em}{0.2em}
\multirow{2}{*}{Resp} &Yes, there are [choice] [type]. I would recommend [name]. It is\\
&at [address] and has [price] entrance. Would that work for you?\\
\specialrule{0em}{0.2em}{0.2em}
User$_{gen}$ &No, but I would like the postcode, address, and phone number.\\
\specialrule{0em}{0.2em}{0.2em}
\multirow{2}{*}{Resp$_{gen}$} &[name]  is located at [address], postcode [postcode],\\
&and the phone number is [phone] .\\
\hline
User &Yes, that will work great. Can I get their phone number please?\\
\specialrule{0em}{0.2em}{0.2em}
\multirow{2}{*}{Resp} &Certainly. [name]'s phone number is [phone]. Is there anything\\
&else i can help you with today?\\
\specialrule{0em}{0.2em}{0.2em}
User$_{gen}$ &Thank you so much for your help.\\
\specialrule{0em}{0.2em}{0.2em}
Resp$_{gen}$ &Thank you for using our system today!\\
\hline
User &No, that is all i need. Thanks!\\
\specialrule{0em}{0.2em}{0.2em}
Resp &You are welcome. Have a great day. Bye.\\
\specialrule{0em}{0.2em}{0.2em}
User$_{gen}$ &\\
\specialrule{0em}{0.2em}{0.2em}
Resp$_{gen}$ &\\
\bottomrule[1.2pt]
\end{tabular}
}
\caption{Interaction case between \Model{}-\modelname{} and \Model{}-US-RL. The generated dialog ends at the third turn.}
\label{tab:interaction-case}
\end{table}

\end{document}